\documentclass[conference]{IEEEtran}
\IEEEoverridecommandlockouts
\usepackage{cite}
\usepackage{amsmath,amssymb,amsfonts}
\usepackage{algorithmic}
\usepackage{graphicx}
\usepackage{textcomp}
\usepackage{xcolor}
\def\BibTeX{{\rm B\kern-.05em{\sc i\kern-.025em b}\kern-.08em
    T\kern-.1667em\lower.7ex\hbox{E}\kern-.125emX}}
\begin{document}

\title{Feature Extraction and Analysis for GPT-Generated Text
}

\author{\IEEEauthorblockN{Aydın Selvioğlu}
\IEEEauthorblockA{\textit{Department of Computer Engineering} \\
\textit{TED University}\\
Ankara, T\"{u}rkiye\\
aydin.selvioglu@tedu.edu.tr}
\and
\IEEEauthorblockN{Venera Adanova}
\IEEEauthorblockA{\textit{Department of Computer Engineering} \\
\textit{TED University}\\
Ankara, T\"{u}rkiye\\
venera.adanova@tedu.edu.tr}
\and
\IEEEauthorblockN{Maksat Atagoziev}
\IEEEauthorblockA{\textit{Department of Software Engineering}\\
\textit{OSTIM Technical University}\\
Ankara, T\"{u}rkiye\\
maksat.atagoziev@ostimteknik.edu.tr}
}

\maketitle

\begin{abstract}
With the rise of advanced natural language models like GPT, distinguishing between human-written and GPT-generated text has become increasingly challenging and crucial across various domains, including academia. The long-standing issue of plagiarism has grown more pressing, now compounded by concerns about the authenticity of information, as it is not always clear whether the presented facts are genuine or fabricated. In this paper, we present a comprehensive study of feature extraction and analysis for differentiating between human-written and GPT-generated text. By applying machine learning classifiers to these extracted features, we evaluate the significance of each feature in detection. Our results demonstrate that human and GPT-generated texts exhibit distinct writing styles, which can be effectively captured by our features. Given sufficiently long text, the two can be differentiated with high accuracy.
\end{abstract}


\section{Introduction}
The first quarter of the \(21^\text{st}\) century witnessed the rise of artificial intelligence (AI) as a revolutionary innovation. Especially the introduction of competent AI chat tools such as ChatGPT and Gemini to our daily lives convinced the masses all around the world about the revolutionary effects of the AI on the future of humanity. The amazing efficiency and productivity enabled by the AI make people support the AI-related developments on the one hand, but, on the other, the versatile, prospective, unforeseen costs of the AI cause deep suspicion about the effects of the AI on the everyday routines of the societies as well as politics and legal regulations. According to \cite{HaenleinK2019}, the possible complications that may be caused by AI can be in three levels: micro-, meso-, macro-levels. In micro level, for instance, the deep learning’s being a black box is a crucial issue. At the meso level, the development of AI could cause unemployment for white-collar employees. Macro level involves the challenges that come with the use of AI for undermining the democratic and peaceful political and social life. In combination, all these levels lead to ethical, legal and philosophical challenges that need to be addressed. Especially vast use of generative AI models causes intense manipulations that could have legal consequences, such as manipulated political elections, plagiarism in academic life, misguiding juridical and medical reports prepared by the models. In particular, critical areas of modern life, such as health, law, education, manufacturing, finance, retail, must deal with the trade-off between cost reduction (efficiency increase) and ethical misuse of AI \cite{KhanBLKWNA2021}. The issues that have legal and ethical aspects concerning the AI-generated products are mainly those of image and text generating. Reports, news, theses, dissertations, assignments, movie scripts, scenarios, all types of visual creations are being subject to conflicts in the sense that they are 'genuine' or not. 

The ability to distinguish between human and AI generated text is significantly difficult for human readers, especially for non experts. Continuous improvement of generative models makes even  the best automatic detectors ineffective \cite{KashnitskyHWTFL2022}. With the advancements, the text generated by these models become even more sophisticated. As reported by~\cite{HerboldHHKT2023}, essays generated by ChatGPT were rated by human experts to be of higher quality than those written by humans. Concerned about this situation,  \cite{GrimaldiE2023} state that if people start using AI content generators as assistants, the writing styles would become homogeneous, losing originality.

There are different experimental results about the ability of AI-generated text detection by humans in the literature. According to \cite{MaLYCHLL2023, LiuHZZSYCW2024} AI-generated text can be detected by human readers with 76\% accuracy. In the work of \cite{GaoHMDRLP2023} 50 medical abstracts were generated using AI-tools, and compared to 50 human-authored abstracts in term of plagiarism scores and detectability by human reviewers. The reviewers could identify 68\% of the generated abstracts correctly. Non-experts' ability to differentiate human-authored text from GPT2/GPT3-generated text was estimated by~\cite{ClarkASHGS2021}. It turned out that non-experts' performance is the same as that of a random guess. Furthermore, after some training the evaluators were only able to improve their accuracy up to 55\%. The study by \cite{IppolitoDCE2020} suggests that the human raters, even the expert ones, have worse classification accuracy than the automatic discriminators. Their results show that the agreement between a pair of randomly selected raters does not exceed 59\% on average. Yet, at the same time ~\cite{BerberSardinha2024} claims that AI-generated text can be identified relatively easy, and that AI is still not capable of accurate reproduction of entangled natural language patterns.

The detection of computer - generated text was a complex task long before the emergence of Large Language Models (LLMs). Thus, \cite{Labbe2012,Labbe2021} showed that computer-generated papers, that were generated by SCIgen, a software which randomly combines strings of words to produce computer science papers, went undetected and published even by the reputable publishers like Springer and IEEE. The authors explain this by the fact that the conferences include a wide range of areas, hence the manuscripts are not always reviewed by the experts in the given field. In another study, ~\cite{Labbe2021_2} noticed that in different research papers the common terms were re-phrased by unusual ones, like using a wording "irregular esteem" instead of "random value", or "haze figuring" instead of "cloud computing", etc. The authors call this strange words as "tortured phrases" that result from automated translation tools or paraphrasing, which are used to conceal plagiarism. 

Aforementioned studies show that human readers are not fully capable of discriminating between human and AI generated text. This is a clear indication that people, even experts, should receive an assistance from automatic detectors. \cite{IppolitoDCE2020}, for instance, suggests that human and detectors make decisions based on different aspects of the text. While humans are good at detecting semantic errors, detectors are good at detecting certain statistical differences in the text. \cite{LiuHZZSYCW2024} analyzed the performance of existing AI content detectors and reported that Originality.ai and ZeroGPT can accurately detect AI generated text. Another study with the same research question was conducted by \cite{WeberwulfABFGPSW2023}, where the authors reveal that the existing detectors tend to classify the text as human written, giving contradictory results. AI content detectors are also biased against non-native speakers. In their research \cite{LiangYMWZ2023} reported that while the essays of native English speakers were correctly classified, more than half non-native speakers' essays were misclassified.

Instead of experimenting with existing detectors, the common approach is to train detectors from scratch \cite{TheocharopoulosATGTP2023,DesaireCIJH2023,GuoZWJNDYW2023,CrothersJVB2022,RodriguezHGSS2022}. However, given the black-box nature of deep models, it is very difficult to understand the decisions made by them. In order to overcome this limitation, several studies \cite{MunozOrtisGV2024, DesaireCIJH2023} extract meaningful features from both human and GPT generated text. These features are typically statistical measurements extracted from the text. Here, we will follow the latter approach. 

In this work, we aim to extract meaningful features from the text data and analyze how they differ in human-written and GPT-generated text. The importance of each feature during classification phase is analyzed further. The data, upon which our experiments are conducted,  consists of theses written in English by non-native speakers. The theses are taken from seven different branches, leading to diverse scientific text data.  

\section{Data}

We collected doctoral dissertations and master theses written in English by Turkish graduate students from the national theses database managed by the Presidency of the Council of Higher Education (Y\"OK) in Turkey. In order to eliminate the possibility that the students got assistance from ChatGPT, we only considered the theses written before 2021. Overall, 84 doctoral dissertations and 6 master theses were selected from seven different branches: physics, biology, industrial engineering, computer engineering, sociology, philosophy, and economics.  We extracted abstracts and introductions from the original theses. The exact thesis titles were used to generate abstracts and introductions using GPT- 3.5. We have observed that the generated introductions are far shorter than the original ones. Hence, we decided to cut the original introductions to comparable lengths. However, we wanted to keep the semantic integrity of the text and not cut the text in the middle of the paragraph. As a result, the number of paragraphs for all human-written introductions was reduced to five. 

We also removed citations from the human-written text, since they might influence the features that we compute subsequently, such as sentence length, word length, etc.

\section{Method}

The main objective of our study is to extract different features from the text data and analyze their differences in two classes: GPT-generated and human-written text. By using extracted features we then aim to identify the most informative ones, that is the features that we believe to have high predictive power. Overall $11$ features were extracted by using statistical, morphological, syntactic, and semantic  measurements.

\begin{figure*}[!ht]
	\centering
	\begin{minipage}[t]{0.35\textwidth}
		\centering
		\includegraphics[width=0.9\textwidth]{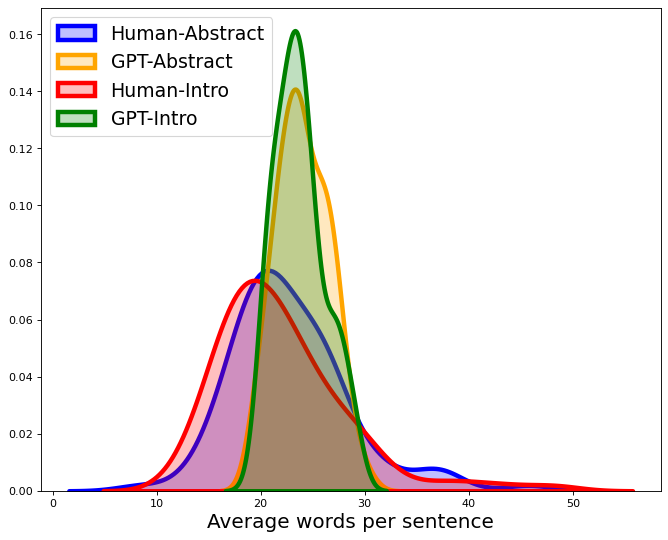}
		\label{fig:titleSim}
	\end{minipage}
	\hspace{-0.15\textwidth} 
	\begin{minipage}[b]{0.55\textwidth}
		\centering
		\raisebox{0.3 cm}{\includegraphics[width=0.9\textwidth]{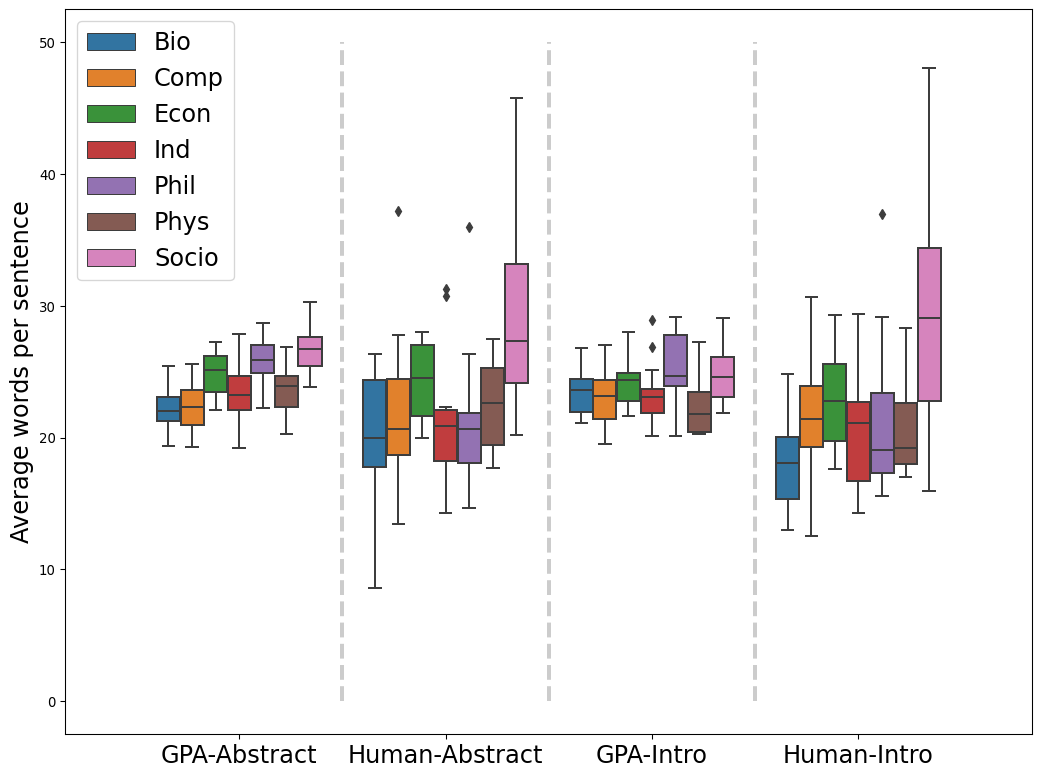}}
		\label{fig:paragraphSim}
	\end{minipage}
	\caption{Average sentence length results. (left) Density plot of sentence length for GPT and human abstracts and introductions. (right) Branch based box plots for sentence length.}
	\label{fig:sentenceSize}
\end{figure*}
\subsection{Statistical Measurements}

Below is the list of statistical measurements derived from the thesis abstracts and introductions:
\begin{itemize}

\item Paragraph size: average number of sentences in paragraphs.

\item Sentence length: average number of words in sentences.

\item Word size: average number of characters in words.

\item Percentage of words consisting of more than 5 characters: we also believe this feature to be useful, as students who are not native English speakers might prefer shorter words. It is given in terms of number of words with more than five characters divided by the total number of words in a text. Considering average would be incorrect because some text have more words than the others.

\item Punctuation marks: count of comas, colons and semicolons. Given as the average number of punctuation marks used in sentences.

\item Entropy: is a measure of unpredictability or complexity in a text. It quantifies the amount of information or randomness in the distribution of elements (e.g., words, characters) within the text. We computed it using $scipy$ library.
\end{itemize}
\subsection{Prefixes and Relative Clauses}

For morphological analysis, we considered the number of words with prefixes. Prefixes are affixes placed at the beginning of a word to modify its meaning. They can change the base word in various ways, such as negating it, intensifying it, or altering its meaning significantly. For example, 'un-' (unhappy), 're-' (rewrite), and 'im-' (impossible). This metric is computed as the number of words with prefixes divided by the total number of words.

We are also interested in how often humans and GPT use relative clauses. Relative clauses enrich sentences by providing additional details and clarifications, making communication more precise and informative. Examples include words like 'who,' 'when,' and 'where.' This metric is computed as the percentage of relative clauses used in a text.

\subsection{Semantic Similarity Measurements}

When GPT generated abstracts and introductions, only the title of a thesis was provided as a source of information. Naturally, we expect the GPT-generated text to be semantically very close to the title. Additionally, we expect that the paragraphs do not diverge from each other,  revolving around the same topic. To capture this, two additional features were introduced: the average similarity of the paragraphs in the text to the title and the average similarity between the paragraphs.

Vector representations of paragraphs and titles are obtained using Sentence-BERT (SBERT) \cite{reimers-2019-sentence-bert}, which is specifically designed to efficiently generate sentence or paragraph embeddings. This is in contrast to BERT \cite{Devlin2019BERTPO}, which was originally designed for token-level tasks. The embeddings from SBERT are optimized for capturing semantic similarity, making them more suitable for tasks that require understanding the overall meaning of a paragraph. Once the embeddings are obtained, the similarities are computed using cosine similarity.

\subsection{Lexical Diversity Measurement}

To measure the lexical diversity of both human-written and GPT-generated text, we use the Measure of Textual Lexical Diversity (MTLD) metric. This metric is designed to assess the variety and range of unique words used within a text and is less sensitive to text length compared to simpler measures like the Type-Token Ratio (TTR). A high MTLD score indicates that the text uses a wide range of vocabulary with less repetitions, suggesting richness of vocabulary. Conversely, a low MTLD score indicates more repetition of words and lower diversity. We used the $lexical-diversity$ library for MTLD measurements.

\section{Results}

All $11$ features described above are computed for every human/GPT generated abstract and introduction. For each of these features we give the density plots. Density plots capture the average distribution without discriminating between branches. In cases when some of the branches behave differently than what we observe on average, we also plot branch-based box plots.
\begin{figure*}[htbp]
	\centering
	\begin{minipage}[t]{0.35\textwidth}
		\centering
		\includegraphics[width=0.9\textwidth]{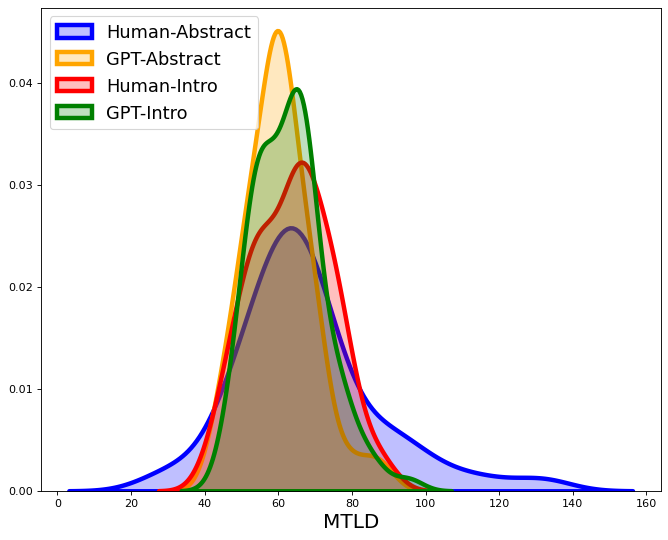}
	\end{minipage}
	\hspace{-0.13\textwidth} 
	\begin{minipage}[b]{0.55\textwidth}
		\centering
		\raisebox{0.2 cm}{\includegraphics[width=0.9\textwidth]{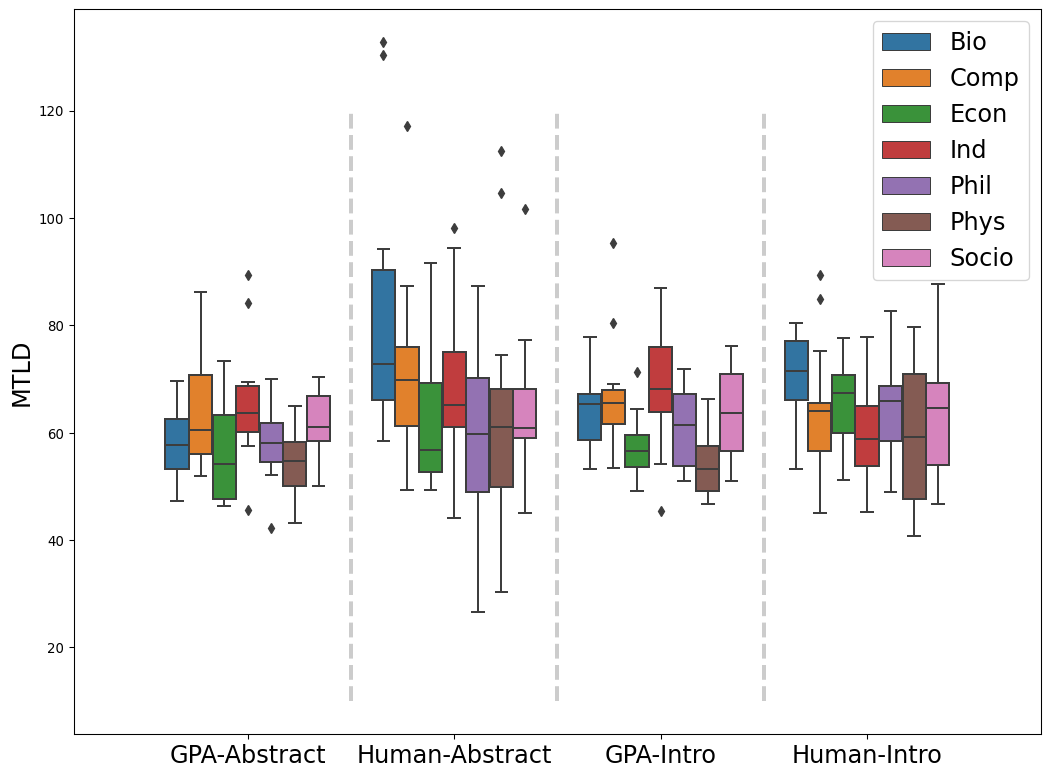}}
	\end{minipage}
	\caption{MTLD results. (left) Density plot of MTLD for GPT and human abstracts and introductions. (right) Branch based box plots for MTLD.}
	\label{fig:mtld}
\end{figure*}
\subsection{Sentence Length}

Sentence length in GPT-generated text is generally longer than in human written text as can be seen in Fig.~\ref{fig:sentenceSize}(left). It seems like GPT has a pattern for word generation since it has a narrow curve which indicates that the sentence length does not deviate much from the average value. The curves for the human-written text have much wider shape and long right tail indicating that the sentence length has wider spread. While GPT has on average longer sentence length, when considering branch-based results (see Fig.~\ref{fig:sentenceSize}(right)) we can see that human-written sociology texts have on average longer sentences. This leads us to a conclusion that considering just the sentence length in the decision process would be misleading as it might depend on the branch.

\subsection{Measure of Textual Lexical Diversity and Entropy}

The results for lexical diversity for both human and GPT-generated text are illustrated in Fig.\ref{fig:mtld}(left). We do not observe differences in the complexity of the text for both human and GPT, except for the human abstract cases, which has long right tail indicating many outliers. More detailed results can be observed in the box plots given per branch in Fig.\ref{fig:mtld}(right). Observe that the human abstract boxes are longer than others, yet the median values are at approximately the same levels as the GPT abstract and introductions for different branches.

\begin{figure*}[htbp]
	\centering
	\begin{minipage}[t]{0.35\textwidth}
		\centering
		\includegraphics[width=0.9\textwidth]{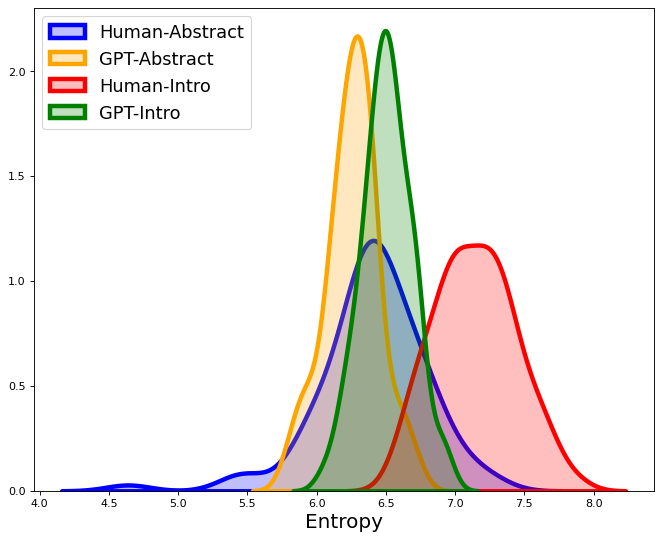}
	\end{minipage}
	\hspace{-0.11\textwidth} 
	\begin{minipage}[b]{0.55\textwidth}
		\centering
		\raisebox{0.2 cm}{\includegraphics[width=0.9\textwidth]{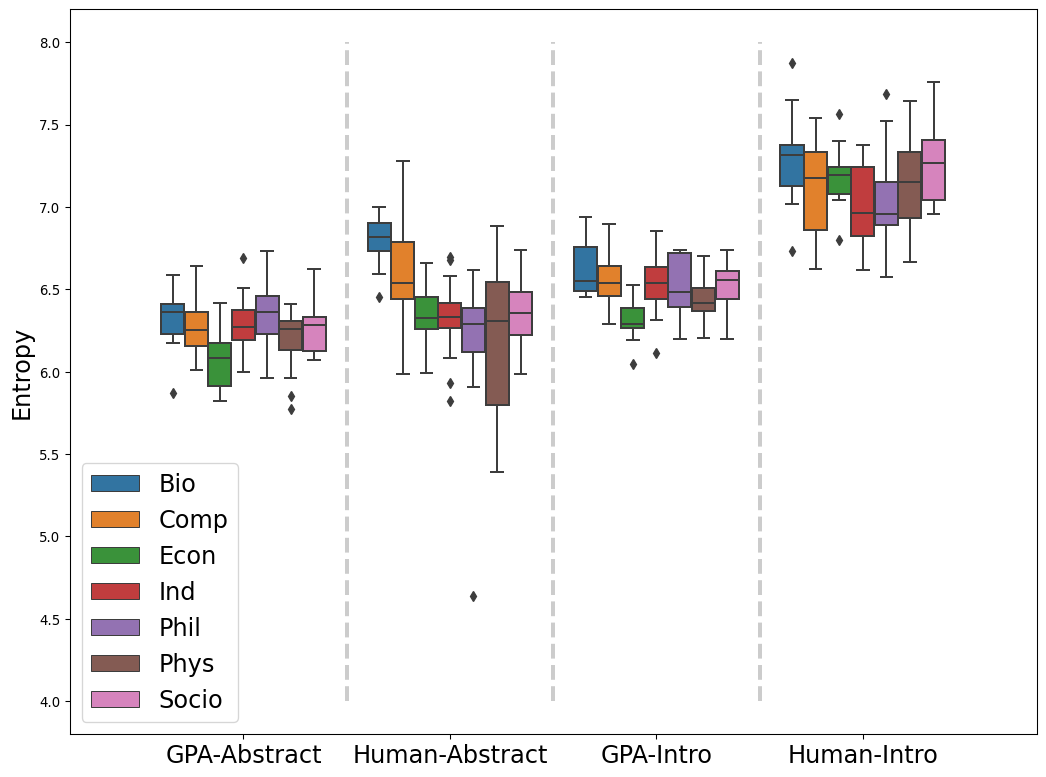}}
	\end{minipage}
	\caption{Entropy results. (left) Density plot of entropy for GPT and abstracts and introductions. (right) Branch based box plots for entropy.}
	\label{fig:Entropy}
\end{figure*}

While the variety of words, as measured by MTLD, is similar for both GPT and human-written text, the same cannot be said about the complexity or distribution of words throughout the text, as measured by entropy. Fig.\ref{fig:Entropy} reports on the results of average entropy. Observe that the human-written introductions, which are much longer than the abstracts, have on average significantly higher entropy. This suggests that the human text is less predictable, with less repetitions. While the lexical diversity of GPT-generated text is similar to human-written text, it has more repetitive or predictable structure of word usage.

\begin{figure}[!htbp]
	\centering
	 \includegraphics[width=0.35\textwidth]{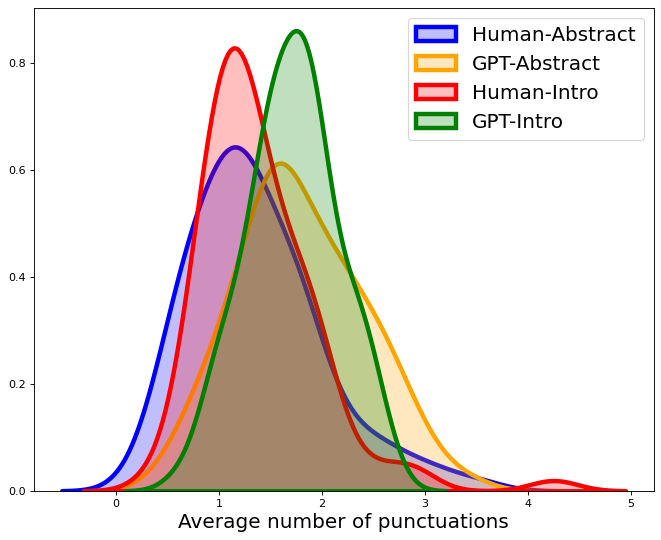}
		
	\caption{Density plot of average number of punctuation in sentences in for GPT and human abstracts and introductions.}
	\label{fig:punct}
\end{figure}
\subsection{Paragraph Size and Punctuation}

Since GPT sentences are, on average, longer, they naturally have more punctuation marks (see Fig.\ref{fig:punct}). This holds true for both abstracts and introductions. Another reason could be that non-native English speakers are not as proficient in punctuation usage as GPT, which was trained on a vast amount of English text and has learned the rules accordingly.

\begin{figure}[!htbp]
	\centering
	 \includegraphics[width=0.35\textwidth]{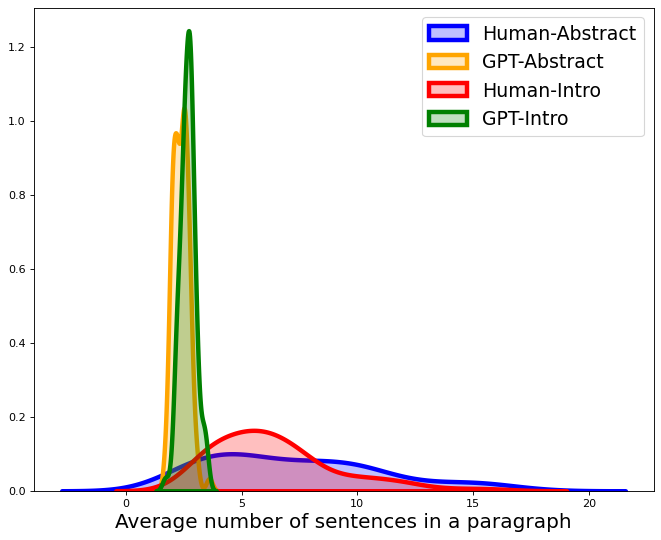}
		
	\caption{Density plot of average number of sentences in paragraphs for GPT and human abstracts and introductions.}
	\label{fig:paraph}
\end{figure}

While GPT generates text with longer sentences, its paragraphs are much shorter compared to human-written text. The density plots showing this are illustrated in Fig.\ref{fig:paraph}. Notice that the GPT curves are narrow for both abstracts and introductions, centered around 2-3 sentences per paragraph. This trend is consistent across all branches.

\begin{figure}[!htbp]
	\centering
	 \includegraphics[width=0.35\textwidth]{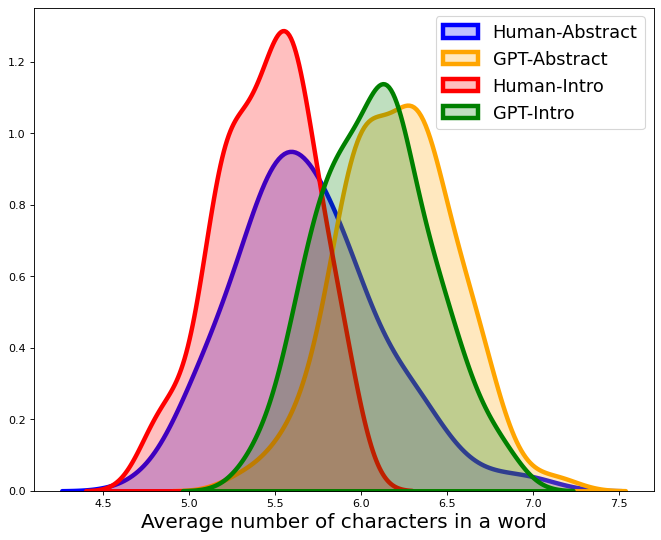}
		
	\caption{Density plot of average number of characters in words for GPT and human abstracts and introductions.}
	\label{fig:wordSize}
\end{figure}

\subsection{Word Size}

\begin{figure}[htbp]
	\centering

		\includegraphics[width=0.35\textwidth]{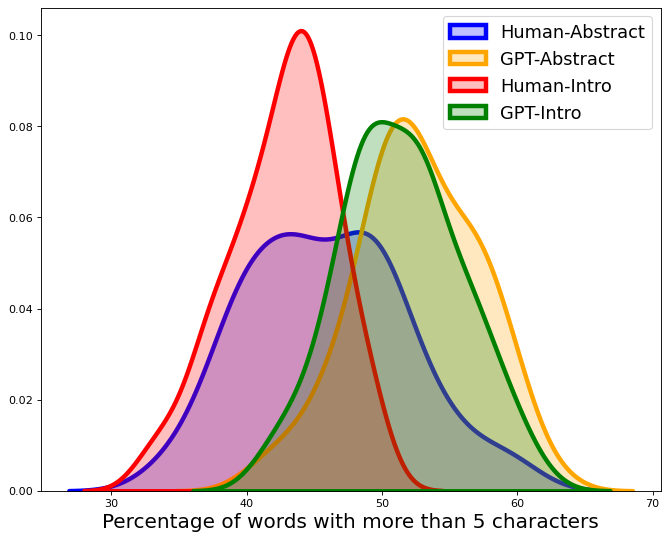}
	\caption{Density plot of percentage of words containing more than five characters for GPT and human abstracts and introductions.}
	\label{fig:wordSize5}
\end{figure}

GPT tends to use longer words than humans, as can be seen in Fig.~\ref{fig:wordSize}. On average the word length is around 6.3 characters, while for human text it is around 5.5 characters. When we consider the percentage of words with more than five characters, unsurprisingly on average more than 50\% of words produced by GPT have length more than five characters as shown in Fig.\ref{fig:wordSize5}.  Since thesis abstracts are the short descriptions of the overall problem and findings, it seems like humans tend to put more effort in finding the right words to describe the problem using fewer words. Notice that the human written abstracts generally contain longer words than the introductions.

\begin{figure}[htbp]
	\centering

		\includegraphics[width=0.35\textwidth]{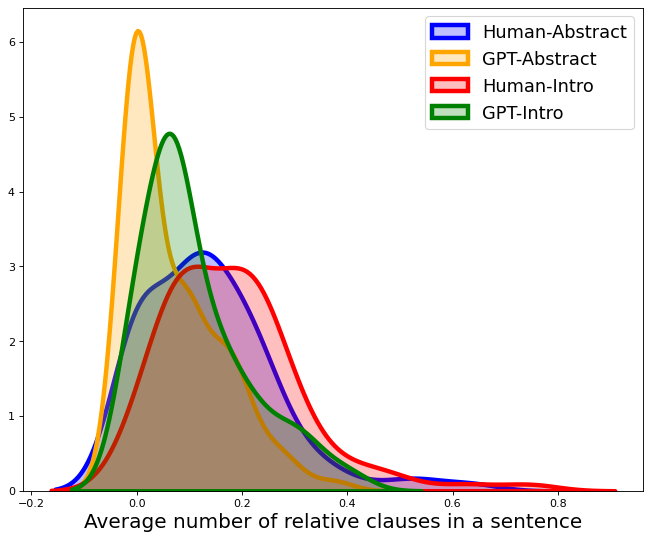}
	\caption{Density plot of average number of relative clauses in text for GPT and human abstracts and introductions.}
	\label{fig:clause}
\end{figure}

\subsection{Prefixes and Relative Clauses}

\begin{figure}[htbp]
	\centering

		\includegraphics[width=0.35\textwidth]{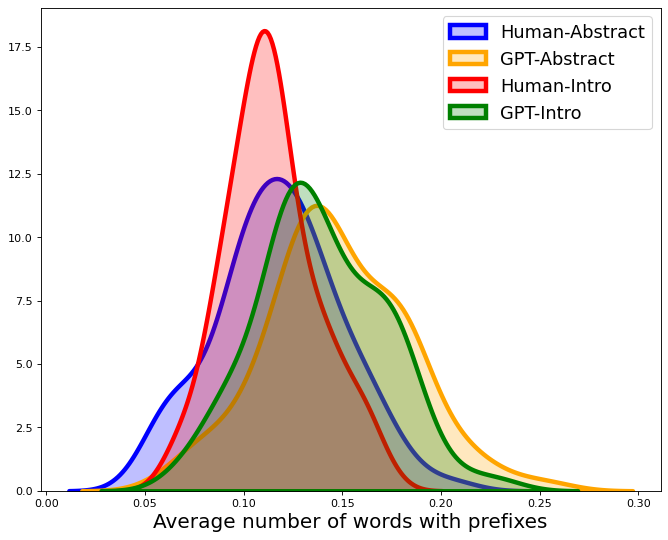}
	\caption{Density plot of average number of words with prefixes GPT and human abstracts and introductions.}
	\label{fig:prefix}
\end{figure}

The average number of relative clauses used in a text is slightly higher for human written text, as illustrated in Fig.\ref{fig:clause}. This result suggests that humans tend to give more details and clarifications in the text. Also, notice that GPT's usage of words with prefixes is slightly more than in human text, as illustrated in Fig.\ref{fig:prefix}. Fewer number of prefixes in the text might indicate that the writers might have chosen simpler language, prioritizing clarity and simplicity. Of course, it might also indicate that the writers have limited vocabulary.

\subsection{Paragraph-Paragraph and Paragraph-Title Similarities}

\begin{figure}[htbp]
	\centering

		\includegraphics[width=0.35\textwidth]{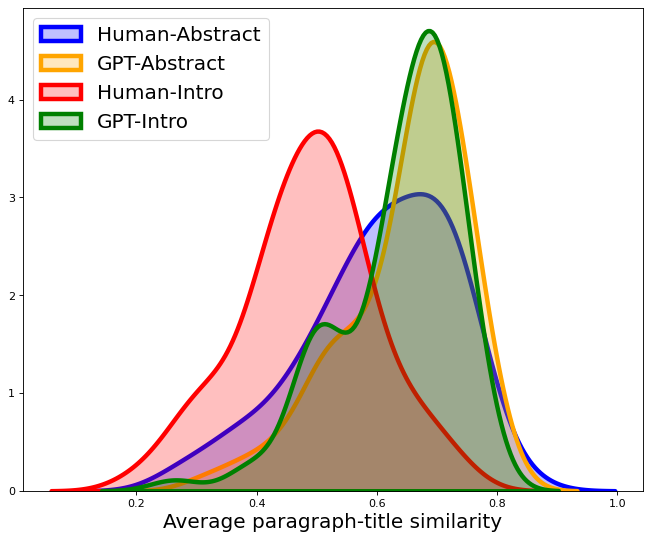}
	\caption{Density plot of average paragraph-to-title similarity for GPT and human abstracts and introductions.}
	\label{fig:title_sim}
\end{figure}

The paragraphs of GPT-generated abstracts and introductions have on average higher similarity to the provided thesis titles, as illustrated in Fig.\ref{fig:title_sim}. This outcome is expected, since the texts were generated from the title context. Since abstracts are short descriptions of the thesis works slightly extending the title, human abstracts also have high contextual similarity to titles. 

\begin{figure}[htbp]
	\centering

		\includegraphics[width=0.35\textwidth]{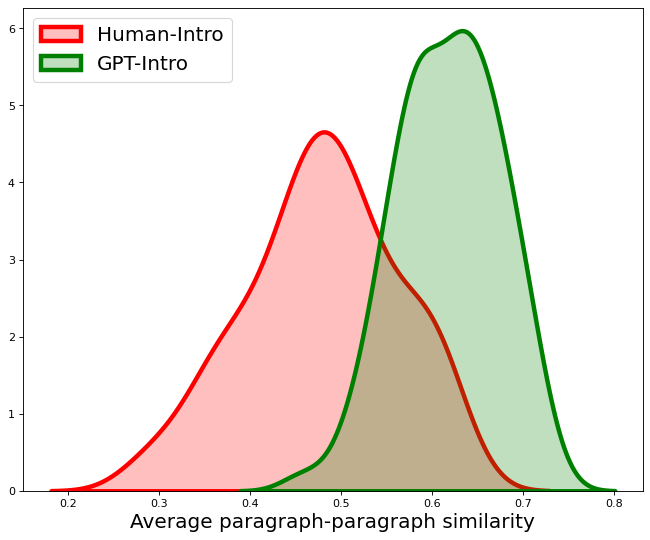}
	\caption{Density plot of average paragraph-to-paragraph similarity for GPT and human abstracts and introductions.}
	\label{fig:paraph_sim}
\end{figure}

The other feature of our interest is the diversity of the paragraphs in the texts. Since the abstracts are generally short, and consist of a single paragraph, we do not include them to this measurement. The density plot of average paragraph-to-paragraph similarity for human and GPT-generated introductions are illustrated in Fig.\ref{fig:paraph_sim}. The average paragraph similarity for GPT-generated introductions are fairly high, indicating that the paragraphs contain contextually close contents.

\subsection{Clustering Results}

\begin{figure*}[htbp]
	\centering
	\begin{minipage}[t]{0.4\textwidth}
		\centering
		\includegraphics[width=0.9\textwidth]{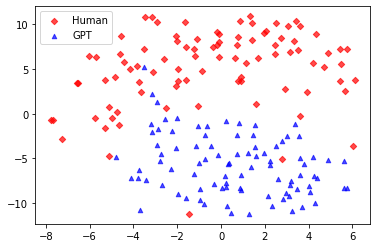}
	\end{minipage}
	\hspace{0.02\textwidth} 
	\begin{minipage}[t]{0.4\textwidth}
		\centering
		\includegraphics[width=0.9\textwidth]{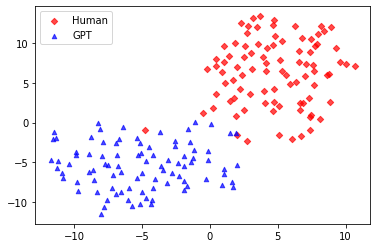}
	\end{minipage}
	\caption{(a) t-SNE results for abstracts. (b) t-SNE results for introductions.}
	\label{fig:tsne}
\end{figure*}

Given the extracted features for our text data, we proceed further with t-Stochastic Neighborhood Embedding (t-SNE) technique \cite{Maaten2008} to visualize how well the chosen features define the clusters. Fig.~\ref{fig:tsne} shows the clustering results for both abstracts and introductions. Human-written abstracts still carry scattered nature, with many outliers. Introductions form two distinct clusters. The results suggest that one still needs a longer text to confidently differentiate between two types of text.

\begin{figure}[htbp]
	\centering
		\includegraphics[width=0.35\textwidth]{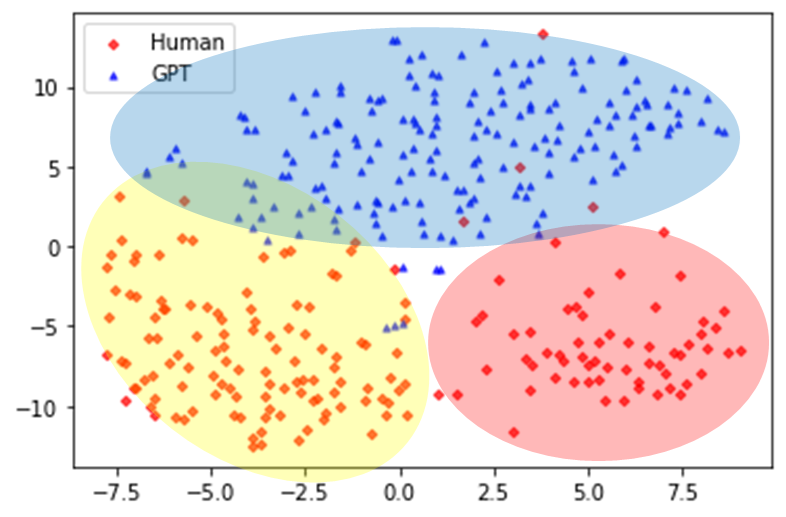}
		
	\caption{t-SNE results for abstracts. Human-written abstracts (encircled in red) separates further from human-written introductions (encircled in yellow) forming different cluster.}
	\label{fig:tsneFull}
\end{figure}

When combining abstracts and introduction, the t-SNE results show three clusters, as demonstrated in Fig.~\ref{fig:tsneFull}. The abstracts and introductions of GPT-generated text are interleaved and form a single cluster. However, human-written text (encircled in red) stands aside from both human-written introductions (encircled in yellow) and GPT-generated text.

\subsection{Feature Importance}

In order to assess the importance of the extracted features in the classification, we first train the Random Forest classifier on our data and then feed the trained model to SHAP (SHapley Additive exPlanations) \cite{SHAP}, which is a method used to explain the predictions of machine learning models by attributing the contribution of each feature to the final prediction. The SHAP values can be positive, indicating that the feature pushed the prediction towards the positive class, and negative, indicating that the feature pushed the prediction towards the negative class. Note that in our case GPT generated text is a positive class. Three models are trained with three different data: abstracts only, introduction only, and combination of abstracts and introductions. As abstracts data has no paragraph to paragraph similarity, this feature was discarded when considering abstracts and combined data. For all cases 30\% of the data is kept for testing. The accuracy results on test data are reported in Table \ref{tab:rf_accuracy}. Note that, when trained separately the abstracts and introductions data have good performance, when combined the accuracy drops drastically.

\begin{table}[h!]
\centering
\caption{Classification accuracy on test data.}
      \begin{tabular}{cc}
        \hline
           & Accuracy \\ \hline
        Abstracts & 98\%\\
        Introductions & 100\%\\
        Combined & 93\%\\ \hline
      \end{tabular}
      \label{tab:rf_accuracy}
\end{table}

\begin{figure}[htbp]
	\centering
	\begin{minipage}[t]{0.45\textwidth}
		\centering
		\includegraphics[width=\textwidth]{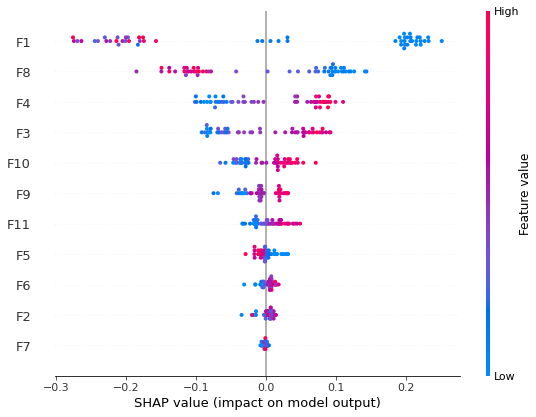}
	\end{minipage}
    \hfill
    \begin{minipage}[t]{0.35\textwidth}
		\centering
		\raisebox{0.7 cm}{\includegraphics[width=\textwidth]{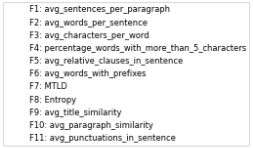}}
	\end{minipage}
	\caption{SHAP feature importance results in introductions data.}
	\label{fig:shapIntro}
\end{figure}

\begin{figure*}[htbp]
	\centering
	\begin{minipage}[t]{0.45\textwidth}
		\centering
		\includegraphics[width=\textwidth]{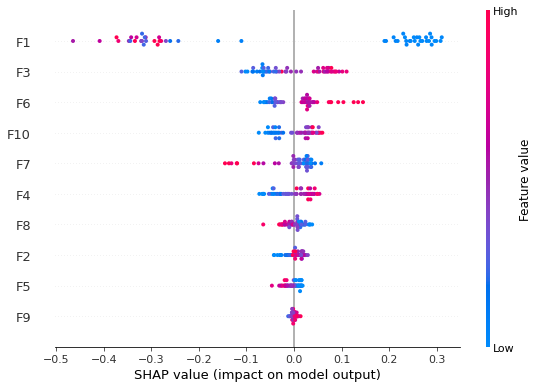}
	\end{minipage}
	\hfill
	\begin{minipage}[t]{0.45\textwidth}
		\centering
		\includegraphics[width=\textwidth]{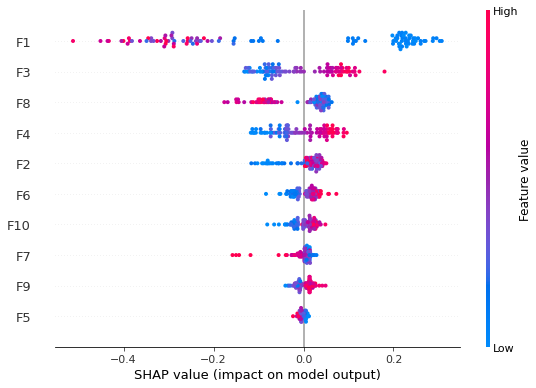}
	\end{minipage}
    \hfill
    \begin{minipage}[t]{0.4\textwidth}
		\centering
		\raisebox{0 cm}{\includegraphics[width=\textwidth]{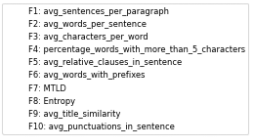}}
	\end{minipage}
	\caption{SHAP feature importance results.(a) Feature importance in abstracts data. (b) Feature importance when abstracts and introductions combined.}
	\label{fig:shap}
\end{figure*}

The summary plots illustrating the importance of features during classification for all three models are given in Fig.~\ref{fig:shapIntro} and \ref{fig:shap}. Y-axis in the plot represents the feature names in the order of importance from top to bottom, and the x-axis represents the SHAP value. The color of each point on the graph encodes the value of the corresponding feature, with red indicating high values and blue indicating low values. Each point is an instance from the original dataset. Observe that paragraph size is the most important feature for all three models. The red dots indicate that human-written text has longer paragraphs. While for abstracts word size is the next important feature, for introductions entropy values come before features representing word size. The semantic similarity of text to title and paragraph to paragraph similarity are one of the most significant features for classification of introduction data. However, in abstracts and combined data title similarity looks like to be of the least significance. MTLD has the least importance value in introductions, while for abstracts it is one of the informative ones. 

\subsection{Paragraph-based Classification}

We continue our analysis with paragraph-based classification using a language model, specifically BERT. In this phase, the input to our model is the text itself, rather than the previously extracted features. The abstracts and introductions are combined and divided into paragraphs, with each paragraph forming a single instance in our dataset. We have 582 paragraphs for abstracts and 948 for introductions. Hence, our dataset consists of 1530 instances. Our aim is to investigate whether the language model can effectively distinguish between human and GPT-generated text based on small excerpts. 
The general approach during text classification is to perform some basic cleaning on text, such as discarding stop words and punctuation. We keep our text intact since we believe that they keep essential information about the human and AI writing styles. The dataset is divided into three: 70\% for training, 15\% for validation and 15\% for testing. We fine-tune BERT for sequence classification for 2 epochs with batch size of 8. BERT tokenizer $max\_length$ parameter was set to 500, which means that if the number of tokens is less than the designated maximum length then it is padded with a special character. 

Due to stochastic nature of BERT several trainings give different results. However, on average 98\% accuracy is achieved on the test set. We observe that every time the misclassified instances are the ones that were human-written, i.e. human-written text is classified as GPT-generated. Another observation is that these misclassified paragraphs are far shorter than the usual human-written paragraphs. This suggests that BERT also considers the paragraph size as an important feature. 

Language models do not consider words separately when learning to classify, but rather learn the relationship between words. Due to black-box nature of deep models it is hard to explain the reasons of certain decisions. Yet, we use SHAP once again to see which words affect the predictions. After SHAP explains the model with the test data, we notice that the words influencing the positive class are on average longer than the words influencing the negative class. Table~\ref{tab:top_words} illustrates top 10 words that end up with high SHAP value for each class. Observe that the positive class also has "?" in the list. It seems like GPT uses question mark a lot when generating text.

\begin{table}[h!]
\centering
\caption{Top 10 words influencing classes.}
      \begin{tabular}{cc}
        \hline
            Positive Class & Negative Class \\ \hline
        although & object\\
        discuss & analyzing\\
        investigating & above\\ 
        interesting & disc\\
        resources & rim \\
        ?  & motivation\\
        demand & class\\
        settings & environments\\
        portfolio & strips\\
        passion & via \\ \hline
      \end{tabular}
      \label{tab:top_words}
\end{table}

\section{Discussion and Conclusion}

From the analysis we can conclude that GPT generates longer sentences, however the distribution of human-written sentences has high standard deviation. This contradicts the findings by \cite{MunozOrtisGV2024}, who report that human sentences are on average longer. However, we also observed that sentence length depends on the field, as human written sociology texts have on average longer sentences than GPT-generated ones. Hence, we conclude that sentence length cannot be used as the only feature during text analysis.

\cite{MaLYCHLL2023} reported that the role of the average word length is significant during AI generated content predictions. Our analysis of feature importance also show that it is one of the most important features during classification. We further observed that BERT pays attention to word sizes during training.

Paragraph size is the most significant feature among all others, and it has the highest influence in predictions. This finding aligns with those reported by  \cite{DesaireCIJH2023}. GPT usually generates paragraphs consisting of 2 or 3 sentences, while humans tend to write much longer paragraphs. BERT misclassified human-written paragraphs as GPT-generated if the size was considerably shorter than normal human-written paragraphs. Based on this fact, we conclude that BERT pays attention to the differences in paragraph sizes.

GPT-generated text has more punctuation symbols and we do not observe any difference in MTLD values. These findings contradict the results of \cite{MunozOrtisGV2024, HerboldHHKT2023}, where they reported that human-written text has a richer vocabulary and uses punctuation symbols more frequently. However, we also observed that human-written text is much more unpredictable according to entropy results.

The semantic similarity of paragraphs to a title and paragraphs to paragraphs proved to be another important feature that we could use in predictions.

As we have stated previously, the detection of AI generated text cannot solely rely on machine learning models and the extracted features. These models and features can only be used to guide a reader through the text, indicating suspicious regions or features. Still, we believe that the final judgment should be made by a human reader.

\bibliographystyle{IEEEtran}

\end{document}